\pdfoutput=1
\documentclass[11pt]{article}

\usepackage[final]{ACL2023}

\usepackage{times}
\usepackage{latexsym}

\usepackage[T1]{fontenc}
\usepackage[utf8]{inputenc}

\usepackage{microtype}

\usepackage{inconsolata}

\usepackage{algorithm}
\usepackage{algorithmic}
\usepackage{amsfonts}
\usepackage{amsmath}
\usepackage{booktabs}
\usepackage{multirow}
\usepackage{graphicx}
\usepackage{enumitem}
\usepackage{textcomp}
\usepackage{booktabs}
\usepackage{subfigure}
\usepackage{array}
\usepackage{bm}
\usepackage{tabularx}
\usepackage{tabulary}
\usepackage{array}
\usepackage{color,ulem}

\title{MaskMoE: Boosting Token-Level Learning \\ via Routing Mask in Mixture-of-Experts}

\author{
  Zhenpeng Su\textsuperscript{\rm 1,2}\quad Zijia Lin\textsuperscript{\rm 3}\quad Xue Bai\textsuperscript{\rm 4}\quad Xing Wu\textsuperscript{\rm 1,2}\quad Yizhe Xiong\textsuperscript{\rm 3}\quad \textbf{Haoran Lian}\textsuperscript{\rm 5}\\ \textbf{Guangyuan Ma}\textsuperscript{\rm 1,2}, \textbf{Hui Chen}\textsuperscript{\rm 3}\quad \textbf{Guiguang Ding}\textsuperscript{\rm 3}\quad \textbf{Wei Zhou}\textsuperscript{\rm 1,2}\quad \textbf{Songlin Hu}\textsuperscript{\rm 1,2}\footnotemark[1] \\
  \vspace{-0.3em}
  \small \textsuperscript{\rm 1}Institute of Information Engineering, Chinese Academy of Sciences\\
  \vspace{-0.3em}
  \small \textsuperscript{\rm 2}School of Cyber Security, University of Chinese Academy of Sciences\\
  \vspace{-0.3em}
  \small \textsuperscript{\rm 3}Tsinghua University,
  \small \textsuperscript{\rm 4}University of Science and Technology of China,
  \small \textsuperscript{\rm 5}Beihang University \\ 
  \vspace{-0.3em}
  \tt\small \texttt{\{suzhenpeng,wuxing,maguangyuan,zhouwei,husonglin\}@iie.ac.cn}\\
  \vspace{-0.3em}
  \tt\small \texttt{byshev333@gmail.com, linzijia07@tsinghua.org.cn, lianhaoran@buaa.edu.cn}\\
  \vspace{-0.3em}
  \tt\small \texttt{huichen@mail.tsinghua.edu.cn, dinggg@tsinghua.edu.cn, xiongyizhe2001@gmail.com}\\
}

\begin{document}
\maketitle
\renewcommand{\thefootnote}{\fnsymbol{footnote}} %将脚注符号设置为fnsymbol类型，即特殊符号表示
\footnotetext[1]{Corresponding authors.} %对应脚注[1]
\renewcommand{\thefootnote}{\arabic{footnote}}
\begin{abstract}
Scaling the size of a model enhances its capabilities but significantly increases computation complexity. Mixture-of-Experts models (MoE) address the issue by allowing model size to scale up without substantially increasing training or inference costs. In MoE, there is an important module called the router, which is used to distribute each token to the experts. Currently, the mainstream routing methods include dynamic routing and fixed routing.
Despite their promising results, MoE models encounter several challenges. Primarily, for dynamic routing methods, the dispersion of training tokens across multiple experts can lead to underfitting, particularly for infrequent tokens. Additionally, though fixed routing methods can mitigate that issue, they compromise on the diversity of representations.
In this paper, we propose \textbf{MaskMoE}, a method designed to enhance token-level learning by employing a routing \textbf{mask}ing technique within the \textbf{M}ixture-\textbf{o}f-\textbf{E}xperts model. MaskMoE is capable of maintaining representation diversity while achieving more comprehensive training.
Experimental results demonstrate that our method outperforms previous dominant Mixture-of-Experts models in terms of both perplexity (PPL) and downstream task performance.
\end{abstract}
% \vspace{-0.3cm}
\section{Introduction}
Large language models have achieved promising performance on various downstream natural language tasks~\cite{DBLP:journals/corr/abs-2302-13971,DBLP:conf/acl/Dai0MZSCW22,DBLP:conf/nips/BrownMRSKDNSSAA20,DBLP:journals/corr/abs-2312-11805,DBLP:journals/corr/abs-2204-02311,radford2019language,DBLP:journals/corr/abs-2112-11446,DBLP:conf/icml/BidermanSABOHKP23}. 
% However, training large language models on extensive textual data has significantly increased computational costs compared to previous works (e.g., BERT~\cite{DBLP:conf/naacl/DevlinCLT19}, ELMO~\cite{DBLP:conf/naacl/PetersNIGCLZ18}, and LSTM~\cite{hochreiter1997long}).
Moreover, according to the scaling law ~\cite{DBLP:journals/corr/abs-2001-08361,DBLP:journals/corr/abs-2203-15556}, as the model size increases, the model's capabilities will continue to grow. However, for dense language models, the computational costs of continuing to scale up are excessive.
In order to further scale up models within computational budgets, sparse activation networks~\cite{DBLP:journals/corr/abs-1904-10509,DBLP:conf/icml/DuHDTLXKZYFZFBZ22} receive widespread attention due to their ability to significantly reduce computational costs by using only a part of parameters per input. A widely studied approach is Mixture-of-Experts (MoE) ~\cite{DBLP:conf/iclr/LepikhinLXCFHKS21,DBLP:conf/icml/DuHDTLXKZYFZFBZ22,DBLP:journals/corr/abs-2401-06066,DBLP:journals/jmlr/FedusZS22,DBLP:conf/nips/RollerSSW21}, which trains multiple expert layers but selects only a subset to process specific inputs. Compared to dense networks of the same model size, MoE effectively reduces computational costs and achieves comparable results in both PPL and downstream~\cite{DBLP:conf/iclr/LepikhinLXCFHKS21,DBLP:conf/icml/DuHDTLXKZYFZFBZ22,DBLP:journals/corr/abs-2401-06066}.

In MoE models, for each input (e.g., a token in the language models), the router in front of the experts needs to decide which experts to feed it in. The commonly used dynamic routing method in MoE is to select the expert with the top $k$ highest confidence based on a probability distribution output by an intermediate layer with learnable parameters that act as the router.
Previous works~\cite{DBLP:conf/iclr/LepikhinLXCFHKS21,DBLP:conf/icml/DuHDTLXKZYFZFBZ22,DBLP:journals/corr/abs-2401-06066,DBLP:journals/jmlr/FedusZS22} show that MoE models trained based on dynamic routing achieve better performance than dense models with the same amount of training computation. 

% e.g., ``basketball''
\begin{figure*}[h]
    \centering
    \includegraphics[width=0.8\textwidth]{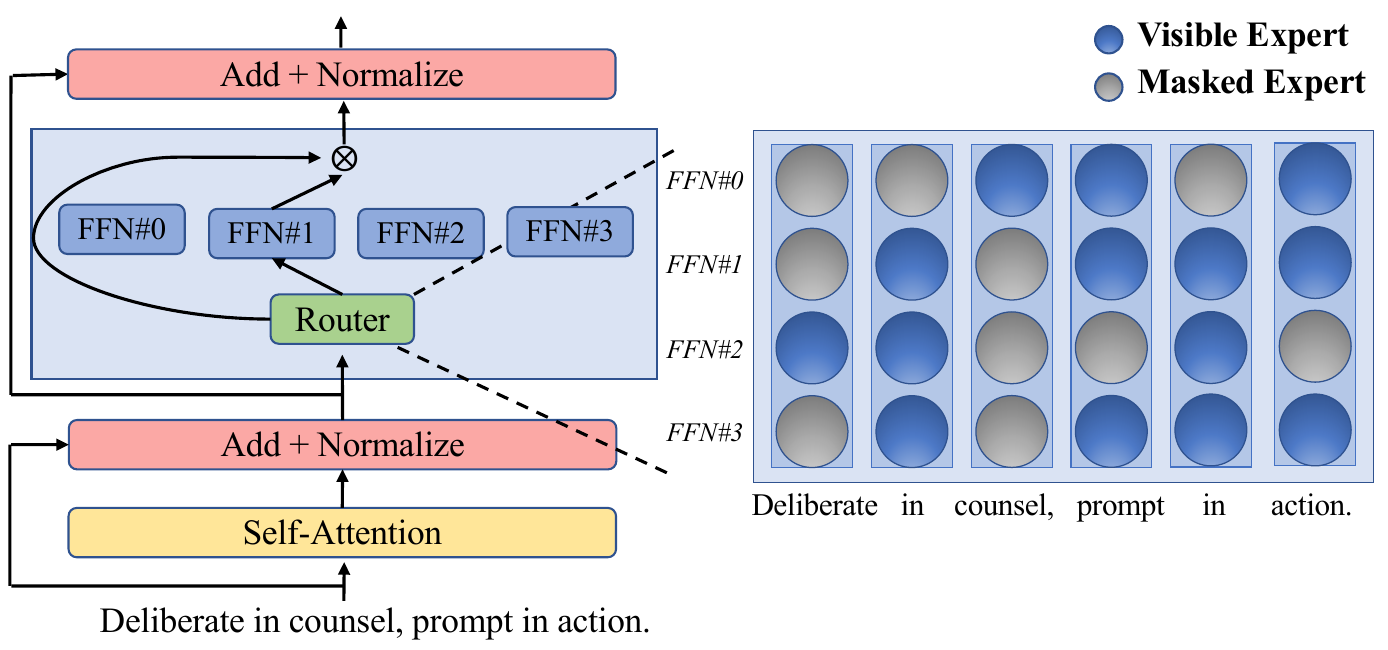}
    % \caption{Illustration of random routing masking method. For each token, some experts will be randomly masked, with only a subset of experts being visible, e.g., for the token ``in'' only expert FFN \#1, \#2, and \#3 are visible.}
    \caption{Illustration of random routing masking method. For each token, some experts will be randomly masked, with only a subset of experts being visible.}
    \label{fig:mask}
\end{figure*}

However, the dynamic routing method introduces a new challenge: the routing fluctuation problem~\cite{DBLP:conf/acl/Dai0MZSCW22}. It implies that as training moves on, the same token is assigned to different experts in different iterations due to the variation of learnable parameters in the router. That has two negative impacts. \textbf{a)} Each expert is trained on only a subset of the occurrences of an identical token, resulting in each expert possibly not learning enough about the token. 
As ~\cite{DBLP:conf/iclr/ShazeerMMDLHD17} has indicated, when there are too many experts, the MoE model exhibits underfitting, and even shows higher PPL than its counterparts with fewer experts. The routing fluctuation problem has relatively less impact on the learning of frequent tokens, as those tokens have sufficient occurrences to ensure that each expert receives adequate training. 
On the contrary, for infrequent tokens, the routing fluctuation problem can disperse each of them across various experts, which probably leads to underfitting for them. 
\textbf{b)} The knowledge each expert learns from the corresponding occurrence subset w.r.t an identical token is hard to share among experts. 
Then the underfitting of experts for some tokens, together with the lack of knowledge sharing among experts, can negatively impact the model’s performance~\cite{zhao2024hypermoe}.
Both issues above hinder scaling the number of experts for an MoE model. Because, as the number of experts increases while keeping the training data unchanged, the average number of token occurrences assigned to each expert further decreases, exacerbating the issues.

To alleviate the issue of routing fluctuations, fixed routing method~\cite{DBLP:conf/nips/RollerSSW21,DBLP:conf/acl/Dai0MZSCW22} is proposed, e.g., Hash Layer.
% Previous fixed routing methods~\cite{DBLP:conf/nips/RollerSSW21} are based on the 
Specifically, Hash Layer~\cite{DBLP:conf/nips/RollerSSW21} pre-assigns each token to a fixed expert. For instance, as shown in Figure \ref{fig:mask}, all occurrences of the token ``deliberate'' can only be routed to the expert FFN\#2.
The fixed routing method to send all the occurrences of a token to the same expert is advantageous for the token's thorough learning, especially for infrequent tokens.
However, fixed routing based on the Hash Layer represents a deficiency in learning representation diversity, especially for those frequent tokens. Previous works~\cite{DBLP:conf/emnlp/HeF0S0T23,DBLP:conf/nips/YangBLN19} show that increasing the number of optional experts for all occurrences of an identical token can help the diversity of learned representations. 
And frequent tokens probably require to be encoded by more experts due to their broader range of usage~\cite{DBLP:conf/cogsci/KorandaZM18}.
For instance, the token ``play'' appears in various contexts and owns different semantic meanings, and thus having a more diverse representation helps to distinguish them.

Therefore, based on the observations above, we need a fixed routing strategy to alleviate the underfitting problem for infrequent tokens, while also requiring more experts for frequent ones to maintain the representational diversity.
To meet the requirements above, in this paper, we propose a routing method called MaskMoE, which includes a \textbf{routing masking} method that adjusts the number of visible experts for tokens of different frequencies, by generating a masking vector for each token in the vocabulary before training.
Specifically, for each infrequent token, the proposed MaskMoE employs routing masking to retain only one visible expert to which the token can be routed, in each MoE layer. For each frequent token, the proposed MaskMoE allow more visible experts for it in each MoE layer.
MaskMoE can enhance token-level learning through such a routing mask. 
Specifically, MaskMoE makes the model learn more intensively about infrequent tokens by routing each to an identical expert in every MoE layer, enabling the expert to be adequately trained w.r.t that token.
Meanwhile, MaskMoE maintains the diversity of representations for frequent tokens by still having multiple experts available for routing, which benefits their representation diversity~\cite{DBLP:conf/emnlp/HeF0S0T23,DBLP:conf/nips/YangBLN19}. Despite frequent tokens being routed to multiple experts, their sufficient quantities can still ensure thorough training. 
% When there are sufficient frequent tokens, appropriately increasing the number of updatable experts can further enhance the model's capabilities. 
Our experimental results indicate that MaskMoE outperforms previous MoE models, in terms of either PPL or downstream task performance.

Our contributions are summarized as follows:

\begin{itemize}[leftmargin=*]
    % \item We highlight that the current dynamic routing method disperses the same tokens to multiple experts, leading to inadequate learning of tokens by some experts, especially infrequent tokens.
    \item We propose MaskMoE, which introduces a routing masking method to assign different numbers of visible experts to tokens based on their frequency to enhance token-level learning. MaskMoE ensures sufficient training for infrequent tokens while maintaining diverse representations for frequent tokens.
    % \item We conduct extensive experiments to analyze the relationship between the number of experts and token frequency. 
    \item We highlight that dynamic routing, which disperses the occurrences of a token into different experts during training, can lead to the underfitting of experts on infrequent tokens, while fixed routing lacks representation diversity for frequent tokens.
    \item  We validate the effectiveness of the proposed MaskMoE with extensive experiments. Experimental results show that it consistently outperforms previous dynamic routing and fixed routing methods for MoE models.
\end{itemize}
% \vspace{-0.3cm}
\section{Related Work}
\subsection{Language Models}
Language models are statistical models designed to optimize the likelihood of token sequences in training data~\cite{DBLP:journals/corr/abs-2302-13971}. 
Initially, language models relied on $n$-gram statistics~\cite{DBLP:journals/pami/BahlJM83,DBLP:journals/tsp/Katz87,DBLP:conf/icassp/KneserN95}. Subsequently, the emphasis shifted to neural network-based models, particularly Recurrent Neural Networks~\cite{DBLP:conf/interspeech/MikolovKBCK10} and their variants like LSTMs~\cite{DBLP:journals/corr/Graves13}. Those models have demonstrated the ability to learn intricate patterns within textual data, achieving significant success in various language modeling applications.
In recent times, Transformers have become the predominant architecture for language models. Notable examples include BERT~\cite{DBLP:conf/naacl/DevlinCLT19}, RoBERTa~\cite{DBLP:journals/corr/abs-1907-11692}, GPT-2~\cite{radford2019language}, UniLM~\cite{DBLP:conf/nips/00040WWLWGZH19}, and T5~\cite{DBLP:journals/jmlr/RaffelSRLNMZLL20}. The introduction of GPT-3~\cite{DBLP:conf/nips/BrownMRSKDNSSAA20}, which boasts 175 billion parameters, marked a significant milestone due to its exceptional performance across numerous downstream tasks. And that then led to a surge in research focusing on large generative language models, including prominent works such as Gopher~\cite{DBLP:journals/corr/abs-2112-11446}, PaLM~\cite{DBLP:journals/corr/abs-2204-02311}, Pythia~\cite{DBLP:conf/icml/BidermanSABOHKP23}, OPT~\cite{zhang2022opt}, GLM~\cite{DBLP:conf/acl/DuQLDQY022,DBLP:conf/iclr/ZengLDWL0YXZXTM23} and LLaMA~\cite{DBLP:journals/corr/abs-2302-13971,DBLP:journals/corr/abs-2307-09288}. Currently, GPT4~\cite{DBLP:journals/corr/abs-2303-08774} achieves truly remarkable results.

However, as the size of the model grows, the computational demands for both training and inference also increase. MoE models achieve scalability by sparsely activating a portion of the model's parameters, allowing for an increase in model size without significantly raising computational overhead. Consequently, MoE models have been receiving increasing attention recently.
\subsection{Mixture-of-Experts}
The concept of Mixture-of-Experts (MoE) models was initially proposed by ~\cite{DBLP:journals/neco/JacobsJNH91}. Later, ~\cite{DBLP:conf/iclr/ShazeerMMDLHD17} applied MoE to LSTM, training an LSTM model with up to 137B parameters. With the rise of the Transformer architecture~\cite{DBLP:conf/nips/VaswaniSPUJGKP17,DBLP:conf/naacl/DevlinCLT19}, Gshard~\cite{DBLP:conf/iclr/LepikhinLXCFHKS21} further applied MoE to Transformers. Subsequently, powerful MoE models such as GLaM~\cite{DBLP:conf/icml/DuHDTLXKZYFZFBZ22} and Switch Transformer~\cite{DBLP:journals/jmlr/FedusZS22} emerged. 

Early works ~\cite{zoph2022st,DBLP:journals/jmlr/FedusZS22,DBLP:conf/icml/DuHDTLXKZYFZFBZ22,DBLP:conf/iclr/LepikhinLXCFHKS21} mainly focused on dynamic routing methods for learning-to-route MoE. Then the Hash Layer ~\cite{DBLP:conf/nips/RollerSSW21} is proposed to use a random hashing method to route tokens to a fixed expert, which even achieved better results than dynamic routing. As an explanation to the advantage of the simple fixed routing method, ~\citeauthor{DBLP:conf/acl/Dai0MZSCW22}~\citeyear{DBLP:conf/acl/Dai0MZSCW22} pointed out that dynamic routing has the problem of routing fluctuation problem, meaning that an identical input is assigned to different experts as training progresses. Routing fluctuation often harms sample efficiency, especially for the learning of infrequent tokens in the language model. On the contrary, other previous works~\cite{{DBLP:conf/emnlp/HeF0S0T23,DBLP:conf/nips/YangBLN19}} also argued that having more experts allows tokens to obtain richer representations, and fixed routing can affect the diversity of representations, especially for frequent tokens. 

Considering both routing fluctuation and representation diversity, we propose MaskMoE, which uses routing masks to alleviate the underfitting problem of infrequent tokens caused by routing fluctuations, while maintaining the representational diversity of frequent tokens.
% \vspace{-0.3cm}
\section{Method}
% \vspace{-0.1cm}
\subsection{Reviewing the Language Models}
Given a tokenized input sequence $\mathbf{x}=({x_{1}, x_{2}, ..., x_{T}})$ consisting of $T$ tokens, a language model generates a probability distribution $\mathbf{p}$ over the vocabulary as output for each token. In nearly all implementations, the Cross-Entropy loss is used as the loss function to maximize the predicted probability $\mathbf{P}_{\cdot, i}^{x_i}$ w.r.t the ground-truth token $x_i$. The training loss $\mathcal{L}_{lm}$ of the generative language model can be formulated as follows:
\begin{equation}
\begin{aligned}
    % \mathcal{L}_{lm} &= - \sum_{i=1}^{T}  \log(\mathbf{p}^{x_{i}}) \\
    \mathcal{L}_{lm} &= - \sum_{i=2}^{T}  \log(\mathbf{P}_{\cdot, i}^{x_i}) \\
    % s.t.,\mathbf{p}_{x_1},\mathbf{p}_{x_2},\ldots,\mathbf{p}_{x_T} 
    % \qquad s.t.,\mathbf{p}_v &= \text{softmax}(W \mathbf{H}^{L}) \\
    s.t., \quad \mathbf{P}_{\cdot, i} &= \text{softmax}(W \mathbf{H}^L_{\cdot, i}) \\
    \mathbf{H}^{L} &= \text{Transformer}(x_{1},x_{2}, \ldots, x_{T})
\end{aligned}
\end{equation}
where $\mathbf{P}_{\cdot, i}$ and $\mathbf{H}_{\cdot, i}^L$ denote the $i$-th column of the matrix $\mathbf{P}$ and $\mathbf{H}^{L}$. 
Here, $\mathbf{H}^L = [\mathbf{h}_{1}^L,\mathbf{h}_{2}^L, \ldots, \mathbf{h}_{T}^L]$ denotes the hidden states of the last layer. 
% With $\mathbf{H}^{L}$, a linear projection layer $W$ is introduced to derive the predicted probability distribution $\mathbf{p}_i$ over the vocabulary where $\mathbf{P} = [\mathbf{p}_1,\mathbf{p}_2,\ldots,\mathbf{p}_T]$.
With $\mathbf{H}^L_{\cdot, i}$, a linear projection layer $W$ is introduced to derive the predicted probability distribution $\mathbf{P}_{\cdot, i}$ over the vocabulary.
The Transformer model consists of $L$ Transformer blocks. Each block is composed of a multi-head self-attention (MHA) module and an FFN module, where the FFN module is generally a two-layer fully connected network. Formally,
\begin{align}
    \mathbf{\hat{h}}_{1}^{l}, \mathbf{\hat{h}}_{2}^{l}, \ldots, \mathbf{\hat{h}}_{T}^{l} = \text{MHA}(\mathbf{h}_{1}^{l-1},\mathbf{h}_{2}^{l-1},\ldots,\mathbf{h}_{T}^{l-1}) \\
    \mathbf{h}_{1}^{l}, \mathbf{h}_{2}^{l}, \ldots, \mathbf{h}_{T}^{l} = \text{FFN}(\mathbf{\hat{h}}_{1}^{l}, \mathbf{\hat{h}}_{2}^{l}, \ldots, \mathbf{\hat{h}}_{T}^{l})
\end{align}
where $l$ represents the $l$-th Transformer block. 
% \vspace{-0.1cm}
\subsection{Reviewing the Mixture-of-Experts}
MoE methods generally replace a single FFN module or multiple/all FFN modules as MoE modules. Each MoE module consists of multiple FFNs, whose outputs are mixed with the routine function $\mathbf{r}(\cdot)$, i.e., the router, as shown in the following formula:
\begin{equation}
\begin{aligned}
    \mathbf{h}_{t}^{l} = \sum_{i}^{N} \mathbf{r}_{i}(\mathbf{\hat{h}}_{t}^{l}) \cdot \text{FFN}_{i}(\mathbf{\hat{h}}_{t}^{l}) 
    \qquad  s.t., \vert \mathbf{r}(\mathbf{\hat{h}}_t^l)\vert_0 \ll N
\end{aligned}
\end{equation}
where $N$ represents the number of experts in a single MoE module, $\mathbf{r}_i$ denotes the routine result w.r.t the $i$-th expert, and $\vert\cdot\vert_0$ denotes the $L_0$-norm. The majority of elements $\mathbf{r}(\cdot)$ are zeros, and thus only a small portion of experts will be activated. Therefore, increasing the total number of experts in MoE does not increase computation or inference time substantially. Regarding the router $\mathbf{r}$ of MoE, it can be divided into fixed routing ~\cite{DBLP:conf/nips/RollerSSW21,DBLP:conf/acl/Dai0MZSCW22} and dynamic routing with learnable parameters ~\cite{DBLP:journals/jmlr/FedusZS22,DBLP:conf/iclr/LepikhinLXCFHKS21,DBLP:journals/corr/abs-2405-04434,DBLP:journals/corr/abs-2401-06066}. The former commonly use random hashing to determine ~\cite{DBLP:conf/nips/RollerSSW21}, before training, which experts each token will be sent to, i.e., $\mathbf{r}$ is unlearnable and preset. In contrast, the latter employs a routing layer with learnable parameters to decide which expert should process an input token, i.e., $\mathbf{r}$ is to be learned during model training.
% \vspace{-0.1cm}
\subsection{Proposed MaskMoE}
As mentioned before, dynamic routing can cause routing fluctuations, while fixed routing, like hash layers, can limit representation diversity.
To mitigate the above two issues in MoE models, we propose MaskMoE to boost token-level learning via routing mask, as illustrated in Figure \ref{fig:mask}. Specifically, for each token, a fixed random masking is applied to the experts, allowing them to be routed only to their corresponding subsets of experts. The formula is represented as follows: 
\begin{align}
\label{eq}
% \mathbf{c} = \{\text{Choice}_i\}_{i=1}^V &\sim \text{Uniform}(\{0, 1, 2, \ldots, N - 1\}) \\
% \mathbf{M} &= [-\infty] \cdot N \\
% \mathbf{M}_i &= 0, \quad \forall i \in \mathbf{C} \\
\mathbf{r} &= \text{softmax}(W_{r} \mathbf{\hat{h}_{t}}^{l} + \mathbf{m}^{t})
\end{align}
where $\mathbf{m}^{t}$ is the masking vector used to control the visibility of token-specific experts for the token $t$. For visible experts, the corresponding elements in $\mathbf{m}^{t}$ are 0, while for invisible experts, the corresponding elements in $\mathbf{m}^{t}$ are $-\infty$. The masking vector is determined before training and does not change during the training process. For multi-layer MoE models, we reuse the same masking vector $\mathbf{m}^{t}$ across different MoE layers.
% $\mathbf{W_{g}}$ is the parameter of the gate, and $e_{i}$ is the expert index that token is sent to.
Here, $W_{r}$ is the learnable parameter of the routing function. It can be seen that we use the preset $\mathbf{m}^t$ to control the visibility of experts like the fixed routing method and meanwhile reserve the learnable parameter $W_r$ as the dynamic routing method, and thus we combine the strengths of both.
The initial masking vector for a token $t$ can be represented by the following formula:
\begin{equation}
\begin{aligned}
\label{eq1}
\mathbf{m}^t &:= -\infty \cdot \mathbf{1}_N \\
\mathbf{C} &= \{\text{C}_i\}_{i=1}^V \sim \mathcal{U}(\{1:N\}) \\
\mathbf{m}_j^t &= 0, \quad \forall j \in \mathbf{C} 
\end{aligned}
\end{equation}
where the number of visible experts is $V$. \(\mathcal{U}(\{1:N\})\) denotes a uniform distribution over the set of integers from \(1\) to \(N\), and $\text{C}_i$ denotes the index of the selected expert for the $i$-th random sampling. It is worth noting that for infrequent tokens, each token usually has only one visible expert, i.e., $V=1$. 
That promotes more thorough learning by the selected expert for such infrequent tokens. For frequent tokens, the number of visible experts is generally greater than 1, i.e., $1<V<=N$, and it is common for $V$ to be less than $N$. We believe that an appropriately sized $V$ not only allows frequent tokens to have higher representation diversity compared to fixed routing but also enables more thorough training for an identical token compared to dynamic routing. 
% The complete process of MaskMoE is illustrated in Algorithm \ref{A1}.
% \begin{algorithm}
% \caption{Proposed MaskMoE Algorithm}
% \label{A1}
% \begin{algorithmic}[1]
% \REQUIRE Training dataset after tokenization $D$, the vocabulary of the tokenizer $T_{v}$, maximal training step number $S$, the number of visible experts $V$, the number of experts $N$
% \STATE Initialize the mask matrix, $\mathbf{M}=\{\mathbf{m}^1, \mathbf{m}^2, \ldots, \mathbf{m}^{T_v}\}$.
% \FOR{each $t$ in $T_{v}$}
%     \STATE $\mathbf{C} = \{\text{C}_i\}_{i=1}^V \sim \mathcal{U}(\{1:N\})$
%     % \STATE $\mathbf{m}^t = \mathbf{M}(t)$ \# The corresponding column w.r.t $t$
%     \STATE $\mathbf{m}_j^t = 0, \quad \forall j \in \mathbf{C}$
% \ENDFOR
% \FOR{$i = 1$ to $S$}
%     \STATE Load an input sequence $\mathbf{x}$ from $D$
%     % \STATE $\mathbf{E} = Embedding(\mathbf{x})$
%     % \STATE \# Load the routing mask for each token in $\mathbf{x}$ \\
%     % \STATE $\mathbf{m} = \mathbf{M}(\mathbf{x})$ 
%     \STATE $\mathbf{m} = \mathbf{m}^t, \quad \forall t \in \mathbf{x}$
%     \STATE Train the MoE models with $\mathbf{x}$ as the input, and add $\mathbf{m}$ to the router according to Equation \ref{eq}
% \ENDFOR
% \end{algorithmic}
% \end{algorithm}

\begin{table*}[!t]
\centering
\scalebox{1}{
\begin{tabular}
{lllcc}
\toprule
Model & Configuration & Params & Activated Params & Pile PPL ($\downarrow$) \\
\midrule
Standard Transformer  & Layers=24, Dense & 468M & 468M & 6.95  \\
\midrule
\multicolumn{5}{l}{\textit{Single-MoE-Layer Setup}} \\
\midrule
SMoE & Layers=24, MoE\_Layers = 1 & 1.3B & 468M & 6.62  \\
Hash Layer & Layers=24, MoE\_Layers = 1 & 1.3B & 468M & 6.56  \\
Share-MoE & Layers=24, MoE\_Layers = 1 & 1.3B & 468M & 6.72  \\
MaskMoE & Layers=24, MoE\_Layers = 1 & 1.3B & 468M & \textbf{6.48} \\
\midrule
\multicolumn{5}{l}{\textit{Multi-MoE-Layer Setup}} \\
\midrule
SMoE & Layers=24, MoE\_Layers = 12 & 10B & 468M & 6.18  \\
Hash Layer & Layers=24, MoE\_Layers = 12 & 10B & 468M & 6.16  \\
Share-MoE & Layers=24, MoE\_Layers = 12 & 10B & 468M & 6.15  \\
MaskMoE & Layers=24, MoE\_Layers = 12 & 10B & 468M & \textbf{6.11} \\
\bottomrule
\end{tabular}
}
\caption{ Perplexity (PPL) results of language modeling.}
\label{ppl_results}
\end{table*}
\subsection{Load Balance Loss}
LLMs are generally trained in a distributed manner. Yet distributed training can lead to load imbalance of MoE models~\cite{DBLP:conf/iclr/LepikhinLXCFHKS21,DBLP:journals/jmlr/FedusZS22,DBLP:conf/acl/Dai0MZSCW22}, where a minority of expert handle the majority of tokens while the others remain idle most of the time. That can negatively impact training efficiency. It is generally desirable for the number of tokens processed by different experts to be roughly equal. To achieve that, a load balancing loss is commonly introduced in the training of MoE models. We follow previous works~\cite{DBLP:journals/corr/abs-2403-07652,DBLP:journals/jmlr/FedusZS22} and adopt a widely used loss as follows:
\begin{equation}
\begin{aligned}
\label{eq
}
\mathcal{L}_{bal} &= N \cdot \sum_{i=1}^{N} w_{i}R_{i} \\
\text{s.t.},\quad w_{i} &= \frac{1}{B} \sum_{j=1}^{B} \mathbb{I} \{argmax(\mathbf{r}^{j})=i\} \\
R_{i} &= \frac{1}{B} \sum_{j=1}^{B} \mathbf{r}_{i}^{j}
\end{aligned}
\end{equation}
where $B$ represents the number of tokens in a mini-batch. $\mathbf{r}^{j}$ denotes the probability distribution of the routing output for the $j$\text{-th} token, derived via Eq. \ref{eq}, and $\mathbf{r}_{i}^{j}$ represents the specific probability value w.r.t the $i$\text{-th} expert. It is noteworthy that for infrequent tokens, which have only one visible expert, the load balancing loss does not apply to them. Since infrequent tokens are assigned to experts through uniform random hashing, the overall load is supposed to be natively balanced, thus having no need for the balancing loss. The balancing loss primarily regulates the routing w.r.t frequent tokens. 
% Therefore, our experiments find that MaskMoE, compared to SMoE, does not experience a significant decrease in training efficiency.

Our final loss is a combination of the language model loss and the load-balance loss:
\begin{align}
    \mathcal{L} = \mathcal{L}_{lm} +  \mathcal{L}_{bal}
\end{align}
% Where $\alpha$ is a tunable hyperparameter.

\section{Experiments}
% \begin{table*}[!t]
% \centering
% \scalebox{1}{
% \begin{tabular}
% {lllcc}
% \toprule
% Model & Configuration & Params & Activated Params & Pile PPL \\

% \midrule
% Standard Transformer  & Layers=24, Dense & 468M & 468M & 6.95  \\
% \midrule
% SMoE & Layers=24, MoE\_Layers = 1 & 1.27B & 468M & 6.62  \\
% Hash Layer & Layers=24, MoE\_Layers = 1 & 1.27B & 468M & 6.56  \\
% Share-MoE & Layers=24, MoE\_Layers = 1 & 1.28B & 468M & 6.72  \\
% MaskMoE & Layers=24, MoE\_Layers = 1 & 1.28B & 468M & \textbf{6.48} \\
% \midrule
% SMoE & Layers=24, MoE\_Layers = 12 & 9.98B & 468M & 6.18  \\
% Hash Layer & Layers=24, MoE\_Layers = 12 & 9.98B & 468M & 6.16  \\
% Share-MoE & Layers=24, MoE\_Layers = 12 & 10.06B & 468M & 6.15  \\
% MaskMoE & Layers=24, MoE\_Layers = 12 & 10.06B & 468M & \textbf{6.11} \\
% \bottomrule
% \end{tabular}
% }
% \caption{ Perplexity results of language modeling. }
% \label{ppl_results}
% \end{table*}

\subsection{Pre-training Dataset}
Following previous works, we use the Pile dataset~\cite{DBLP:journals/corr/abs-2101-00027} as pre-training data. The Pile dataset is a large-scale, publicly available corpus, containing 22 domains and over 825GB of English text data. For our experiments, we use the well-known LLaMA tokenizer for tokenization, with a vocabulary size of $32k$. We follow ~\cite{DBLP:conf/nips/Xie0DDLLLL0Y23, su-etal-2024-mile} to calculate the sampling rate for each domain based on the number of tokens after tokenization. Due to the limited computational budget, and following the pretraining settings of ~\cite{DBLP:conf/nips/Xie0DDLLLL0Y23,su-etal-2024-mile,DBLP:journals/corr/abs-2403-07652,xiong2024temporal,lian2024scaffold}, all models are pre-trained with 100B tokens.

Then, to identify infrequent tokens and frequent tokens, we calculate the frequency of each token in the training set and sort them in descending order of frequency. 
We categorize the top tokens that cover $P \times 100\%$  of the dataset as frequent ones, and the remaining as infrequent ones, where $P$ is a tunable hyperparameter, ranging from 0 to 1. 
% We categorize the top tokens that cover $P\%$  of the dataset as frequent ones, and the remaining $(1-P)\%$ as infrequent ones, where $P$ is a tunable hyperparameter, ranging from 0 to 100. 

\subsection{Compared Models}
We compare the proposed MaskMoE with four remarkable baselines for validation experiments. Following previous works~\cite{DBLP:conf/nips/RollerSSW21,DBLP:journals/jmlr/FedusZS22,DBLP:conf/acl/Dai0MZSCW22}, unless otherwise specified, our experiments select the top-$1$ expert for each MoE layer.
\begin{itemize}[leftmargin=*]
    \item \textbf{Dense} represents a standard Transformer language model.
    \item \textbf{SMoE} denotes a Switch Transformer~\cite{DBLP:journals/jmlr/FedusZS22}, where the router is a learnable layer. Unless otherwise specified, each MoE layer has 64 experts. 
    \item \textbf{Hash Layer}~\cite{DBLP:conf/nips/RollerSSW21} signifies that through a random hash method, each token is assigned to a fixed expert before training. Unless otherwise specified, each MoE layer has 64 experts. 
    \item  \textbf{Share-MoE} is a hybrid dense and MoE model created using residual connections~\cite{DBLP:conf/icml/RajbhandariLYZA22}. Models with shared experts are currently a popular architecture~\cite{DBLP:journals/corr/abs-2401-06066,DBLP:journals/corr/abs-2405-04434,zhao2024hypermoe,DBLP:conf/icml/RajbhandariLYZA22}. Unless otherwise specified, Share-MoE has 1 shared expert and 128 routed experts, where each expert is 0.5$\times$ of the size of a standard FFN. During both training and inference, in addition to the activation of shared experts, the top-$1$ expert is also selected from the pool of 128 experts. 
    In such a setup, Share-MoE has the same number of floating-point operations (FLOPs) and activation parameters as SMoE and Hash Layer. That allows for a fair comparison between Share-MoE, SMoE, and Hash Layer.
    \item  \textbf{MaskMoE} aims to achieve a balance between representation diversity and thorough training, via the proposed routing masking method. Similarly to Share-MoE, MaskMoE also employs the remarkable shared expert architecture, which has been proven by ~\cite{DBLP:journals/corr/abs-2401-06066,DBLP:conf/icml/RajbhandariLYZA22} to be beneficial for the training of MoE models. 
\end{itemize}

\subsection{Experimental Setup}
Following ~\cite{DBLP:journals/corr/abs-2302-13971,DBLP:journals/corr/abs-2307-09288,DBLP:journals/corr/abs-2403-07652}, we adopt the LLaMA architecture with 24 Transformer blocks and a hidden-state dimension of 1024. We employ the AdamW ~\cite{DBLP:conf/iclr/LoshchilovH19} optimizer with a cosine learning rate decay schedule. 
% For the MoE model, following ~\cite{DBLP:conf/icml/RajbhandariLYZA22,DBLP:journals/jmlr/FedusZS22,DBLP:conf/nips/RollerSSW21,DBLP:journals/corr/abs-2401-06066}, each MoE layer has 64 experts.

For the MoE model, following ~\cite{DBLP:conf/nips/RollerSSW21}, we conduct experiments under both single-layer and multi-layer settings. For the single-layer setting, we replace the FFN layer with the MoE layer only for the last Transformer block. 
% In the case of multi-layer setting, we follow the Gshard ~\cite{DBLP:conf/iclr/LepikhinLXCFHKS21} by interleaving MoE layers every other layer. 
For the multi-layer MoE models, we follow Gshard~\cite{DBLP:conf/iclr/LepikhinLXCFHKS21}, and replace the FFN layer with MoE layer for every other Transformer block, resulting in a total of $12$ MoE layers in this setting.
For single-layer MoE models and dense models, we use a learning rate of $3e^{-4}$, while for Multi-layer MoE models, following ~\cite{DBLP:conf/icml/LewisBDGZ21}, we employ a learning rate of $1e^{-4}$ to ensure stable convergence of the models.
% If not specified otherwise, for infrequent tokens, the visible expert count $V$ is set to 8, and for infrequent tokens, the visible expert count $V$ is set to 1. 

If not specified otherwise, the top most frequent tokens that cover 40\% of the training set are considered frequent tokens, while the remaining are considered infrequent tokens, i.e., $P = 0.4$. Empirically, for frequent tokens, the visible expert count $V$ is set to 8, and for infrequent tokens, the visible expert count $V$ is set to 1. All of our implementations are based on the \texttt{DeepSpeed}\footnote{https://github.com/microsoft/DeepSpeed} library~\cite{DBLP:conf/icml/RajbhandariLYZA22,DBLP:conf/kdd/RasleyRRH20}, which offers robust support for MoE-distributed training. By default, we enable the random token selection method~\cite{DBLP:journals/corr/abs-2109-10465} implemented in the library to facilitate faster model convergence and better runtime efficiency. 

\begin{table*}[!t]
\centering
\scalebox{0.75}{
\begin{tabular}{l|p{1.45cm}<{\centering}p{1.45cm}<{\centering}p{1.45cm}<{\centering}p{1.45cm}<{\centering}p{1.45cm}<{\centering}p{1.45cm}<{\centering}p{1.45cm}<{\centering}p{1.45cm}<{\centering}p{1.45cm}<{\centering}}
\toprule
 Model  & BoolQ & Hellaswag & LAMBADA & PIQA & SIQA & StoryCloze & Arc-e & TriviaQA & WebQs   \\
\midrule
Standard Transformer   & 56.02 & 40.73 & 52.55 & 67.62 & 40.79 & 63.55 & 51.43 & 7.44 & 4.97  \\
\midrule
\multicolumn{10}{l}{\textit{Single-MoE-Layer Setup}} \\
\midrule
SMoE & 55.26 &43.66 & 53.46 & 68.28 & 41.15 & 64.14 & 51.98 & 10.00 & \underline{6.99}  \\
Hash Layer  & 56.61 &\underline{43.88} & \underline{54.67} & \textbf{69.53} & \underline{41.76} & \underline{64.19} & \textbf{53.70} & 9.82 & 6.89  \\
% MaskMoE  & \textbf{57.40} & \textbf{54.86} & \underline{69.10} & \textbf{41.97} & \textbf{65.68} & \underline{53.91} & \textbf{10.01} & \underline{6.89}  \\
Share-MoE & \textbf{58.87} &43.00 & 53.00 & 68.39 & 41.45 & \underline{64.19} & 51.92 & \textbf{10.54} & 6.89  \\
MaskMoE & \underline{58.38} &\textbf{44.47} & \textbf{55.36} & \underline{68.99} & \textbf{41.86} & \textbf{65.15} & \underline{53.14} & \underline{10.39} & \textbf{7.14}  \\
\midrule
\multicolumn{9}{l}{\textit{Multi-MoE-Layer Setup}} \\
\midrule
SMoE   & 56.18 & 46.92 & 56.37 & 69.91 & 41.20 & 64.99 & 54.34 & \underline{13.54} & \textbf{7.33}  \\
Hash Layer   & 57.16 & 46.61 & \underline{56.68} & \textbf{71.00} & 40.83 & \underline{65.47} & 55.05 & 12.97 & 5.95  \\
Share-MoE & \textbf{58.71} & 46.94 & 56.39 & 69.69 & 40.84 & 65.42 & \textbf{55.89} & 12.77 & \underline{6.94}  \\
% MaskMoE   & \textbf{58.78} & \textbf{56.82} & \textbf{71.06} & \textbf{41.20} & 65.05 & \underline{55.21} & \textbf{13.75} & \textbf{7.38}  \\
MaskMoE & \underline{58.32} & \textbf{47.46} & \textbf{57.46} & \underline{70.62} & \textbf{41.91} & \textbf{65.69} & \underline{55.35} & \textbf{15.20} & 6.74  \\
\bottomrule
\end{tabular}
}
\caption{Performances of language models on downstream tasks. The best score is marked in \textbf{bold}, and the second best is \underline{underlined}.}
\label{expert_results}
\end{table*}

\subsection{Main Results}
In this section, we first present the model's PPL on the Pile validation set. 
Then, following~\cite{DBLP:journals/corr/abs-2302-13971, DBLP:conf/nips/BrownMRSKDNSSAA20, su-etal-2024-mile,DBLP:journals/corr/abs-2401-06066}, we conduct tests on various downstream benchmarks, including zero-shot tests for BoolQ~\cite{DBLP:conf/naacl/ClarkLCK0T19}, HellaSwag~\cite{DBLP:conf/acl/ZellersHBFC19}, LAMBADA~\cite{DBLP:conf/acl/PapernoKLPBPBBF16}, PIQA~\cite{DBLP:conf/aaai/BiskZLGC20}, SIQA~\cite{DBLP:journals/corr/abs-1904-09728}, StoryCloze~\cite{DBLP:journals/corr/MostafazadehCHP16}, and Arc-e~\cite{DBLP:journals/corr/abs-2102-03315}. Following ~\cite{DBLP:journals/corr/abs-2302-13971,su-etal-2024-mile}, we conduct 5-shot tests for TriviaQA~\cite{DBLP:conf/acl/JoshiCWZ17} and WebQs~\cite{DBLP:conf/emnlp/BerantCFL13}. Among them, TriviaQA and WebQs use \texttt{exact match} as the metric, while the remaining benchmarks are evaluated based on \texttt{accuracy}.
For a fair comparison, we use the open-source evaluation tool \texttt{lm-evaluation-harness}\footnote{https://github.com/EleutherAI/lm-evaluation-harness} for assessment.

\subsubsection{Perplexity Results}
Table \ref{ppl_results} shows the main results of language modeling on the Pile validation sets. 
% MaskMoE outperforms existing MoE methods on the Pile validation set, whether in settings with a single MoE layer or multiple MoE layers. 
With the same number of activated parameters during inference (i.e., 468M), MoE models consistently achieve improvements (i.e., substantially lower PPL) over the dense model. In comparison to SMoE and Hash Layer, under the \texttt{Single-MoE-Layer} setup, MaskMoE has respectively reduced the PPL on the Pile validation set by 0.14 and 0.08. In the \texttt{Multi-MoE-Layer} setup, their corresponding PPL are, respectively, reduced by 0.07 and 0.05. Compared to Share-MoE, MaskMoE also achieves a PPL decrease of 0.24 in \texttt{Single-MoE-Layer} setup and 0.04 in \texttt{Multi-MoE-Layer} setup. Such results demonstrate the effectiveness of our proposed MaskMoE.
 
 It is noteworthy that Share-MoE achieves worse results than SMoE and Hash Layer in the \texttt{Single-MoE-Layer} setup, even with a built-in shared expert. However, MaskMoE consistently maintains superiority over them in both single-layer and multi-layer setups. That indicates MaskMoE exhibits stronger robustness across various settings.

\subsubsection{Benchmark Results}
As shown in Table \ref{expert_results}, we present the performance of the models on downstream tasks. We can observe that MoE models consistently and significantly outperform the dense models in downstream tasks. More importantly, MaskMoE substantially outperforms the other MoE baselines, whether in \texttt{Single-MoE-Layer} setup or \texttt{Multi-MoE-Layer} setup. 

Specifically, Compared to SMoE, MaskMoE consistently outperforms it in all 9 benchmarks with the \texttt{Single-MoE-Layer} setup and 8 of 9 benchmarks with the \texttt{Multi-MoE-Layer} setup. Against Hash Layer, MaskMoE excels in 7 of 9 benchmarks with the \texttt{Single-MoE-Layer} setup and in 8 of 9 benchmarks with the \texttt{Multi-MoE-Layer} setup. Compared to Share-MoE, MaskMoE demonstrates significant improvements in 7 of 9 benchmarks with the \texttt{Single-MoE-Layer} setup and in 6 of 9 benchmarks with the \texttt{Multi-MoE-Layer} setup.

The result above well demonstrates the performance enhancement achieved by using our proposed router masking method that controls the visibility of experts for different tokens. We believe the improvements are due to two factors. Firstly, compared to SMoE and Share-MoE, infrequent tokens are only routed to a single expert, ensuring more thorough training for those tokens. Secondly, in contrast to the Hash Layer, frequent tokens have multiple expert options, which promotes diversity in their representation learning. 

\section{Analyses}
We conduct further experiments to provide more insightful analyses on the proposed MaskMoE. Considering the constraints of computational resources and following previous works~\cite{DBLP:conf/nips/RollerSSW21,DBLP:conf/aaai/XieHCW23,DBLP:conf/acl/Dai0MZSCW22}, unless otherwise specified, most of the analytical experiments are conducted under the setting of a single MoE layer.
\subsection{Impact of Shared Expert}
To explore the impact of shared experts, we conduct ablation experiments on them (i.e., removing the shared experts), and the results are shown in Table \ref{ppl_results_ablation}. 
We separately report the performance on the benchmark and the PPL on the Pile validation set. Note that after removing the shared experts, we adopt the same architecture as SMoE and Hash Layer, where the MoE layer consists of 64 fully-sized FFN. 

We find that even without Shared-Experts, MaskMoE still outperforms SMoE, Hash Layer, and Share-MoE in terms of benchmark scores and PPL, with Table \ref{expert_results} as references. Specifically, compared to SMoE, MaskMoE w/o Shared-Experts excels in 8 of 9 benchmarks. Additionally, compared to Hash Layer and Share-MoE, MaskMoE w/o Shared-Experts gains significant improvements in 7 of 9 benchmarks. Similarly, MaskMoE w/o Shared-Experts, yields lower PPL than SMoE, Hash Layer, and Share-MoE, with Table \ref{ppl_results} as references. Such a result further validates the soundness of our proposed MaskMoE. Moreover, incorporating Shared-Expert structures continuously improves MaskMoE's performance. As shown in Table \ref{ppl_results_ablation}, compared to MaskMoE w/o Shared-Expert, MaskMoE excels in 5 of 9 benchmarks and has a lower PPL in the Pile validation set. We attribute it to that the shared expert structure promotes greater specialization in token representation~\cite{DBLP:journals/corr/abs-2405-04434}.
\begin{table}[!ht]
\centering
\scalebox{0.9}{
\begin{tabular}{lcc}
\toprule  
 & MaskMoE & w/o Shared-Expert \\
\midrule 
    BoolQ & \textbf{58.38}  & 57.40 \\
    Hellaswag & \textbf{44.47}  & 44.10 \\
    LAMBADA & \textbf{55.36}  & 54.86 \\
    PIQA & 68.99  & \textbf{69.10} \\
    SIQA & 41.86  & \textbf{41.97} \\
    StoryCloze & 65.15 & \textbf{65.69} \\
    Arc-e & 53.14  & \textbf{53.91}  \\
    TriviaQA & \textbf{10.39}  & 10.01 \\
    WebQs & \textbf{7.14}  & 6.89 \\
\midrule
    PPL($\downarrow$) & \textbf{6.48}  & 6.51 \\
\midrule
\end{tabular}
}
\caption{Impact of Shared Expert on benchmark performances and the Pile validation PPL. The best score is marked in \textbf{bold}.}
\label{ppl_results_ablation}
\end{table}

\begin{table}[!t]
\centering
\scalebox{0.8}{
\begin{tabular}
{lllcc||lllcc}
\toprule
& $P$ & $V_{a}$ & $V_{b}$ & PPL & $P$ & $V_{a}$ & $V_{b}$ & PPL \\
\midrule
& 0.0 & 8 & 1 & 6.558 $\spadesuit $ & 0.6 & 8 & 1 & 6.518 \\
& 0.2 & 8 & 1 & 6.518 & 0.8 & 8 & 1 & 6.539 \\
& 0.4 & 8 & 1 & \textbf{6.506} & 1.0 & 8 & 1 & 6.566 $\spadesuit $ \\
% & 0.6 & 8 & 1 & 6.518 \\
% & 1.0 & 8 & 1 & 6.566 $\spadesuit $ \\
\bottomrule
\end{tabular}
}
\caption{Impact of the frequency threshold $P$ for the separation of frequent and infrequent tokens. $\spadesuit $ denotes no distinction between frequent and infrequent tokens.}
\label{router_p}
\end{table}

\begin{table}[!t]
\centering
\scalebox{0.9}{
\begin{tabular}
{lllcc||lllcc}
\toprule
& $P$ & $V_{a}$ & $V_{b}$ & PPL & $P$ & $V_{a}$ & $V_{b}$ & PPL\\
\midrule
& / & 64 & 64 & 6.618$\spadesuit$ & 0.4 & 64 & 1 & 6.549 \\
& / & 8 & 8 & 6.566 & 0.4 & 8 & 1 & \textbf{6.506} \\
& / & 4 & 4 & 6.551 & 0.4 & 4 & 1 & 6.511 \\
& / & 1 & 1 & 6.558 $\clubsuit$ & 0.4 & 16 & 1 & 6.509 \\
\bottomrule
\end{tabular}
}
\caption{ Impact of visible experts. $\spadesuit$ denotes SMoE, $\clubsuit$ denotes Hash Layer.}
\label{router_mask}
\end{table}

% We separately report the average scores\footnote{Due to the limited space, here we simply average all metrics on downstream tasks to get an average score for each model.} of the benchmarks and the PPL on the Pile validation set. 
% Note that after removing the shared experts, we adopt the same architecture as SMoE and Hash Layer, where the MoE layer consists of 64 fully-sized FFN.

% We find that even without shared experts, MaskMoE still outperforms SMoE, Hash Layer, and Share-MoE in terms of average benchmark scores and PPL. 
% % This further validates the design of MaskMoE, which ensures sufficient training for tokens, especially infrequent ones, while maintaining diversity for frequent tokens.
% Such a result further validates the soundness of our proposed MaskMoE.
% Moreover, incorporating shared expert structures continuously improves MaskMoE's performance. We attribute it to that the shared expert structure promotes greater specialization in token representation~\cite{DBLP:journals/corr/abs-2405-04434}.

\subsection{Impact of Hyperparameters for Router Masking}

Considering the proposed MaskMoE without shared experts still outperforms the compared MoE methods, to show the impact of router mask more clear and minimize the impact of the shared expert, here we keep the same experimental setting with the shared expert removed and conduct further experiments to analyze the impact of hyperparameters of our proposed MaskMoE.

The performance of MaskMoE is predominantly influenced by two hyperparameters: the boundary threshold (i.e., $P$) for the separation of frequent and infrequent tokens, and the maximum number of visible experts (i.e., $V$) for tokens. 
As shown in Table \ref{router_p} and \ref{router_mask}, we conduct parameter searches for both and report their PPL values on the Pile validation sets. 
Specifically, here $V_{a}$ represents the number of experts visible for frequent tokens, and $V_{b}$ represents the number of experts visible for infrequent tokens. 
% Notably, when $V_{a} = V_{b} = 1$, MaskMoE is equivalent to the Hash Layer, and when $V_{a} = V_{b} = 64$, MaskMoE is equivalent to SMoE.

As shown in Table \ref{router_p}, we report the impact of the boundary threshold $P$. Here, $P=0.0, V_a=8, V_b=1$ and $P=1, V_a=8, V_b=1$ indicate no distinction between frequent and infrequent tokens, with every token having $1$ and $8$ visible experts, respectively. Their performances are significantly worse than those under the settings of $0 < P < 1$, given that $V_a=8, V_b=1$. That indicates tokens of different frequencies indeed require distinct visible experts, validating the effectiveness of our design. Additionally, we observe that regardless of the settings of $P$ where $0 < P < 1$, the PPLs are consistently lower than $P = 0.0$ or $P = 1$. That validates the robustness of MaskMoE.

As shown in Table \ref{router_mask}, we report the impact of the number of visible experts for frequent and infrequent tokens. Firstly, compared to SMoE with $V_a=V_b=64$, concentrating tokens into fewer visible experts through our proposed routing mask, like $V_{a} = V_{b} = 8$ or $V_{a} = V_{b} = 4$, yields better outcomes. It indicates that routing tokens more densely to a reduced number of visible experts can enhance thorough training and may improve model performance. Moreover, we observe that  $V_{a} = 8, V_{b} = 1$ outperforms $V_{a} = V_{b} = 8$. It well verifies that further reducing the number of visible experts for infrequent tokens is even more beneficial. Secondly, it is noteworthy that for MaskMoE, when $V_{b}=1$, regardless of the value of $V_{a}$ (whether it is 4, 8, 16, or 64), the performance of MaskMoE consistently outperforms that of SMoE$(i.e., V_a=V_b=64)$ and Hash Layer$(i.e., V_a=V_b=1)$. That also demonstrates the robustness of MaskMoE. 

\begin{figure}
    \centering
    \subfigure[None-Shared Expert Structures]{
        \includegraphics[width=0.48\textwidth]{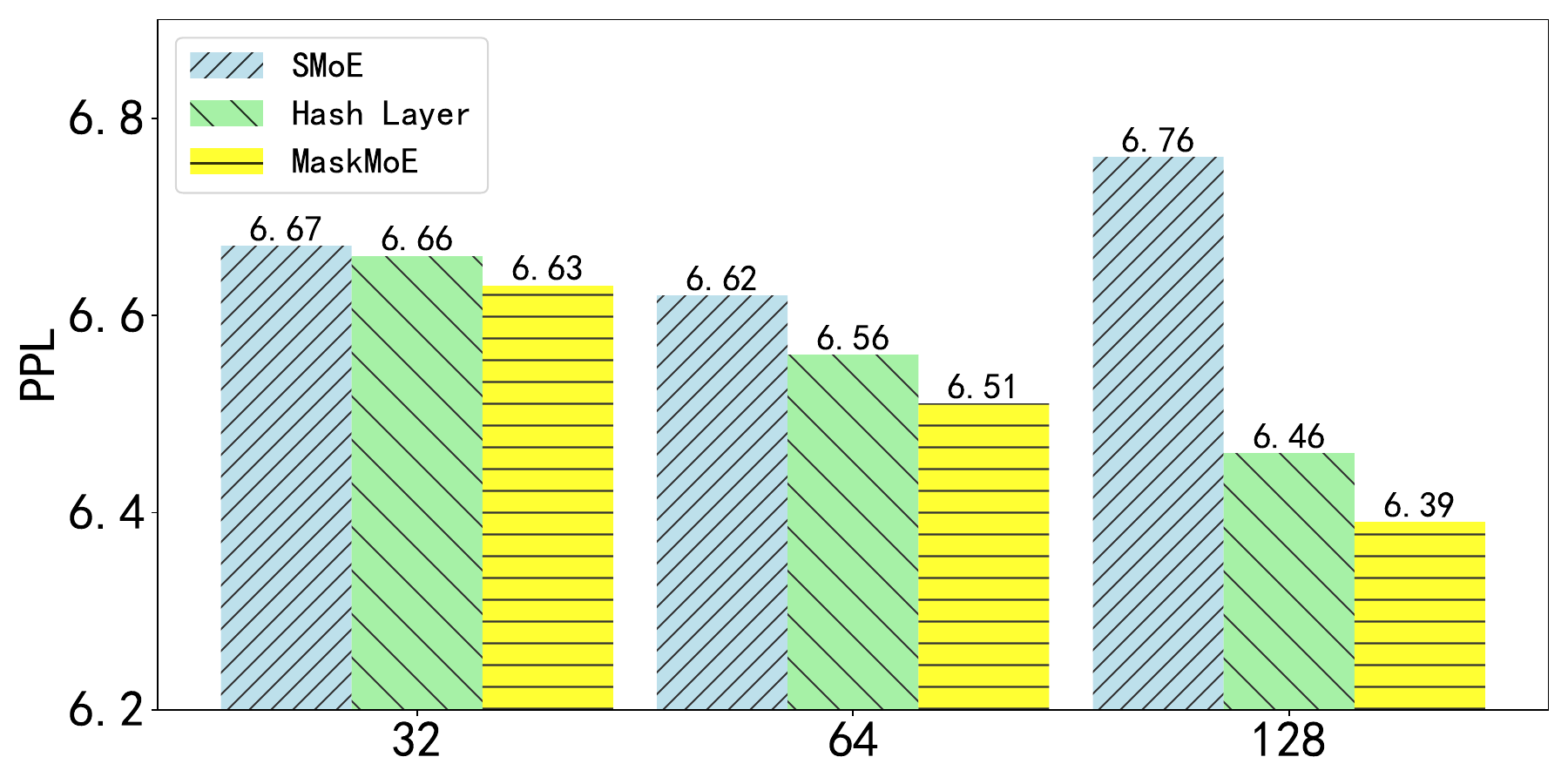}
        \label{fig:subfig1}
    }
    \subfigure[Shared-Expert Structures]{
        \includegraphics[width=0.48\textwidth]{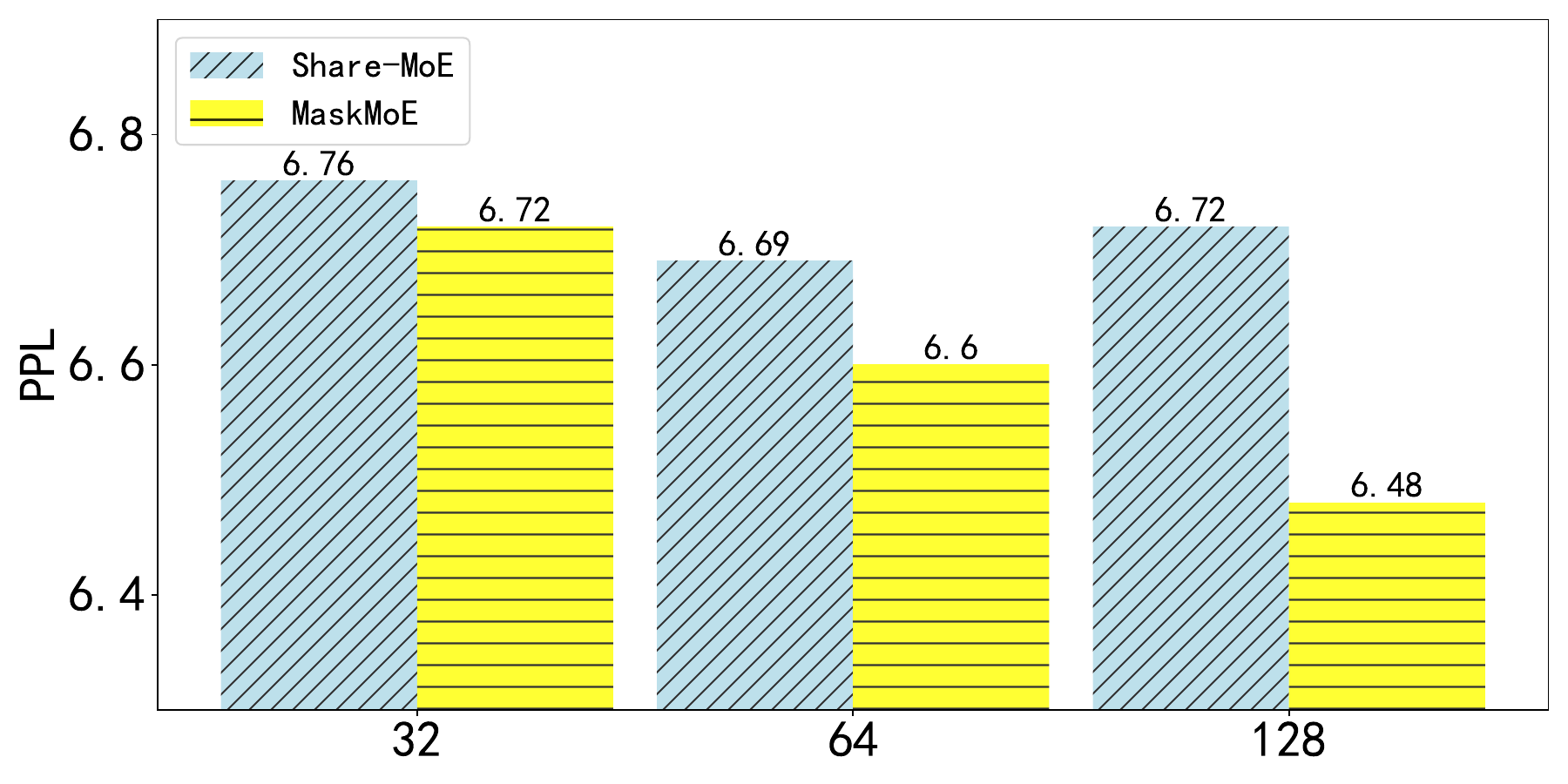}
        \label{fig:subfig2}
    }
    \caption{Comparison of MoE-based Transformers with different numbers of experts.}
    \label{fig:overall_fig}
\end{figure}

\subsection{Impact of the Number of Experts}
To investigate the variation in model performance under different configurations of expert numbers, we conducted experiments with 32, 64, and 128 experts. The experimental results are illustrated in Figure \ref{fig:overall_fig}. 

First of all, we observe that for SMoE and Share-MoE, increasing the number of experts does not necessarily lead to a lower PPL. In fact, when the number of experts reaches 128, both models even exhibit higher PPLs compared to having only 64 experts.
We attribute it to that having more experts leads to the occurrences of an identical token being spread among more experts, and that disables each expert to learn adequately about this token, leading to model underfitting.
In contrast, the performance of Hash Layer and MaskMoE continue to improve as the number of experts increases. Actually, when the number of experts increases, the average number of token occurrences assigned to each expert decrease. However, Hash Layer and MaskMoE can still ensure that each expert is fed with all or a large portion of the occurrences of a token,  though the number of distinct tokens decreases for each expert. That makes the data for each expert more consistent and easier to learn.
% In contrast, the performance of Hash Layer and MaskMoE continue to improve as the number of experts increases. We attribute it to that, despite the growing number of experts, the number of visible experts for an identical token in Hash Layer and MaskMoE remains a fixed and small number, allowing the experts to learn adequately for specific tokens. And thus both seem to be more favorable for scaling the number of experts. 

Secondly, compared to Hash Layer, MaskMoE consistently performs better across all configurations. Although Hash Layer also enables experts to be adequately trained for specific tokens, MaskMoE excels in the diversity of representation learning for frequent tokens, which contributes to MaskMoE's overall advantage over Hash Layer.

Finally, regardless of the number of experts, MaskMoE consistently yields the best performance. 
That further verifies its superiority and reasonableness.

\subsection{Performance Analysis of Top-k Routing}
% \begin{table}[!t]
% \centering
% \scalebox{0.9}{
% \begin{tabular}
% {lc|lc}
% \toprule
% Model  & PPL($\downarrow$) & Model  & PPL($\downarrow$)\\
% \midrule
% SMoE  & 6.55 & Share-MoE  & 6.58 \\
% Hash Layer  & 6.50 & MaskMoE   & \textbf{6.46} \\
% \bottomrule
% \end{tabular}
% }
% \caption{The PPL on the Pile validation sets under the top-$2$ gating. The best score is marked in \textbf{bold}.}
% \label{top2_ppl_results_ablation}
% \end{table}

\begin{table}[!t]
\centering
\scalebox{0.8}{
\begin{tabular}
{lcccc}
\toprule
% SMoE  & 6.55 & Share-MoE  & 6.58 \\
% Hash Layer  & 6.50 & MaskMoE   & \textbf{6.46} \\
Model & SMoE & Hash Layer & Share-MoE & MaskMoE \\
\midrule
PPL($\downarrow$) & 6.55 & 6.50 & 6.58 & \textbf{6.46} \\
\bottomrule
\end{tabular}
}
\caption{The PPL on the Pile validation sets under the top-$2$ gating. The best score is marked in \textbf{bold}.}
\label{top2_ppl_results_ablation}
\end{table}
We further analyze the performance of the MoE models under the top-$k$ ($k=2$) routing mechanism. In the SMoE model, the top-$2$ experts are selected from 64 fully-sized FFN experts based on the router's output scores. For the Hash Layer, two experts are randomly chosen from 64 fully-sized FFN experts using a random hash function. Regarding Share-MoE and MaskMoE models, in addition to the shared experts, the top-$2$ experts are selected from the 128 half-sized FFN experts based on the router's output scores. All settings share the same number of active parameters. Note that in MaskMoE, the number of visible experts is doubled from the original setting, i.e., $V_a=16, V_b=2$. As shown in Table \ref{top2_ppl_results_ablation}, the model's PPLs on the Pile validation sets are presented. MaskMoE consistently yields the best performance, achieving a 0.12 decrease in PPL compared to Share-MoE. That further verifies the effectiveness.
% \vspace{-0.3cm}
\section{Conclusions}
In existing MoE language models, the widely-used dynamic routing methods such as SMoE, exhibit routing fluctuations, and route the occurrences of a token to multiple experts during training, potentially causing the model to encounter underfitting, especially for infrequent tokens. Although fixed routing like Hash Layer can mitigate the routing fluctuation issue, it struggles with capturing rich features for frequent tokens. 
% To tackle the routine fluctuation issue and the representation diversity learning issue in the meantime, we propose a routing masking method termed MaskMoE, aiming at ensuring adequate learning of infrequent tokens while maintaining rich representations learning for frequent tokens. 
Considering both routing fluctuation and representation diversity, we propose MaskMoE, a routing masking method that ensures adequate learning for infrequent tokens and rich representations for frequent tokens. 
Extensive experiments show that MaskMoE outperforms SMoE, Hash Layer, and Share-MoE on various downstream tasks and significantly reduces PPL on the Pile validation sets.

\section{Limitiations}
In this paper, we categorize tokens simply into frequent and infrequent tokens by their frequency. Such a classification method seems somewhat rigid. 
A smoother classification method could potentially yield better results for our proposed MaskMoE.
Due to computational constraints, we do not conduct further experiments. We leave the exploration of smoother partitioning methods for our future work. 

\bibliography{anthology,custom}
\bibliographystyle{acl_natbib}

% \appendix

% \section{Example Appendix}
% \label{sec:appendix}

% This is a section in the appendix.

\end{document}